%% file: neurips_2025.tex
\documentclass{article}

\usepackage[preprint]{neurips_2025}

\usepackage[utf8]{inputenc} 
\usepackage[T1]{fontenc}    
\usepackage{hyperref}       
\usepackage{url}            
\usepackage{booktabs}       
\usepackage{amsfonts}       
\usepackage{nicefrac}       
\usepackage{microtype}      
\usepackage{xcolor}         
\usepackage{soul}

\usepackage{multirow}
\usepackage{graphicx}
\usepackage{amsmath}
\usepackage{amssymb}
\usepackage{comment}

\usepackage{algorithm}
\usepackage{algpseudocode}
\usepackage{booktabs}

\usepackage{lipsum,adjustbox}
\usepackage{todonotes}
\usepackage{pifont}
\usepackage{filecontents}
\usepackage[table]{xcolor}
\usepackage{pgfplots}
\usepackage{subcaption}
\usepackage{pgf-pie} 
\usepackage{tabularx}
\usepackage{makecell}
\usepackage[most]{tcolorbox}
\usepackage{xcolor} 
\usepackage{wrapfig}

\title{WildSci: Advancing Scientific Reasoning from In-the-Wild Literature}

\author{%
  Tengxiao Liu
  \quad \textbf{Deepak Nathani}
  \quad \textbf{Zekun Li}
  \quad \textbf{Kevin Yang}
  \quad \textbf{William Yang Wang} \\
  University of California, Santa Barbara \\
  \texttt{tengxiao@ucsb.edu} \\
}

\makeatletter
\renewcommand{\@noticestring}{}
\makeatother

\begin{document}

\maketitle

\begin{abstract}
Recent progress in large language model (LLM) reasoning has focused on domains like mathematics and coding, where abundant high-quality data and objective evaluation metrics are readily available. In contrast, progress in LLM reasoning models remains limited in scientific domains such as medicine and materials science due to limited dataset coverage and the inherent complexity of open-ended scientific questions. To address these challenges, we introduce WildSci, a new dataset of domain-specific science questions automatically synthesized from peer-reviewed literature, covering 9 scientific disciplines and 26 subdomains. By framing complex scientific reasoning tasks in a multiple-choice format, we enable scalable training with well-defined reward signals. We further apply reinforcement learning to finetune models on these data and analyze the resulting training dynamics, including domain-specific performance changes, response behaviors, and generalization trends. Experiments on a suite of scientific benchmarks demonstrate the effectiveness of our dataset and approach. We release WildSci to enable scalable and sustainable research in scientific reasoning.
\footnote{\url{https://huggingface.co/datasets/JustinTX/WildSci}}
\end{abstract}

\input{ch_intro}

\input{ch_method}

\input{ch_experiments}

\input{ch_results}

\input{ch_conclusion}

\begin{ack}
We gratefully acknowledge support via the UC Santa Barbara NSF Quantum Foundry funded via the Q-AMASE-i program under award DMR-1906325.
\end{ack}

\bibliographystyle{abbrvnat}
\bibliography{custom}


\appendix

\input{ch_appendix}


\newpage

\end{document}

%% file: ch_intro.tex
\section{Introduction}

Advancing AI for science requires models that combine domain-specific expertise with strong reasoning capabilities to support real-world scientific discovery~\citep{DBLP:conf/emnlp/0044CJWJW024, DBLP:journals/corr/abs-2502-14499, DBLP:journals/corr/abs-2502-18864,DBLP:journals/corr/abs-2503-17604}. 
Recent advances in large language model (LLM) reasoning have achieved impressive progress in mathematical and coding domains~\citep{DBLP:journals/corr/abs-2409-12122,DBLP:journals/corr/abs-2402-03300}, especially driven by the rapid development of reinforcement learning techniques for reasoning tasks~\citep{DBLP:journals/corr/abs-2402-03300, DBLP:journals/corr/abs-2501-12599, DBLP:journals/corr/abs-2503-24290, DBLP:journals/corr/abs-2503-14476, DBLP:journals/corr/abs-2503-20783, DBLP:journals/corr/abs-2503-18892}.
These areas offer natural advantages: they involve objective, verifiable answers and have abundant high-quality datasets available in the community. This enables the development of reasoning models with access to scalable, well-structured training and evaluation.

In contrast, scientific domains remain relatively underexplored in the context of reinforcement learning with verifiable rewards (RLVR)~\citep{DBLP:journals/corr/abs-2411-15124, DBLP:journals/corr/abs-2503-23829}. Scientific questions are often more complex and multifaceted, requiring not only logical and mathematical skills but also deep domain-specific knowledge to contextualize and analyze information accurately.
Existing datasets are mostly skewed toward traditional natural sciences such as physics, chemistry, and biology, and tend to be limited in both scope and disciplinary diversity~\citep{DBLP:journals/corr/abs-2311-12022, DBLP:conf/nips/LiHIKG23,DBLP:conf/nips/ZhangHZDYWYD024, DBLP:journals/corr/abs-2501-15587}. As a result, many scientific areas, particularly interdisciplinary fields like materials science and medicine, remain underrepresented~\citep{DBLP:journals/corr/abs-2406-07835}. Furthermore, most existing data sources are drawn from textbooks or general pretraining corpora~\citep{DBLP:conf/nips/YueZZC24,DBLP:journals/corr/abs-2502-13124}, which lack the specificity of research-level content.

To address this gap, we propose leveraging peer-reviewed scientific literature as a rich yet underutilized source for constructing domain-specific science questions in a fully automated pipeline. 
Unlike textbooks or problem sets, scientific papers reflect the depth, rigor, and complexity of real-world research, making them well-suited for advancing models toward research-level reasoning skills. 
This approach offers several key advantages: (1) it grounds questions in real-world applications and expert-validated context; and (2) it enables the creation of new questions that are unlikely to appear in pretraining corpora, helping mitigate issues of data contamination.

Another key challenge arises from the nature of science itself: many science questions are inherently open-ended and do not have a single verifiable answer. 
For example, explaining the observed decline in species richness in Figure~\ref{fig:pipeline} requires scientific judgment including interpreting evidence, reasoning about underlying mechanisms, and constructing plausible explanations. 
To address this, we adopt a more structured formulation by framing scientific reasoning tasks as multiple-choice questions (MCQs). MCQs are widely used in existing science benchmarks and offer a practical format for evaluation~\citep{DBLP:conf/iclr/HendrycksBBZMSS21, DBLP:journals/corr/abs-2311-12022, DBLP:conf/nips/WangMZNCGRAHJLK24, DBLP:journals/corr/abs-2502-14739}. This structure provides clear supervision signals, making it easier to define rewards systematically while preserving the richness of scientific reasoning. This simple setting offers a natural testbed for extending RL advances from mathematical domains to scientific reasoning tasks.

In this work, we develop a generalizable approach for creating training data grounded in real-world scientific research, and to extend RLVR reasoning to scientific domains. Our contributions are summarized as follows.

(1) We introduce a \textbf{fully automated data synthesis pipeline} that generates domain-specific questions from peer-reviewed scientific papers, followed by refinement and model voting  to ensure data quality.

(2) We construct WildSci, a dataset of 56K questions \textbf{spanning 9 scientific disciplines and 26 subdomains}, providing broad and diverse coverage for scientific reasoning.

(3) We provide comprehensive analysis on how WildSci enables \textbf{effective transfer of RLVR method to scientific domains}. Models trained on WildSci show consistent improvements on multiple science benchmarks, including GPQA, SuperGPQA, and MMLU-Pro.

With new papers continuously emerging in the community, WildSci provides a sustainable data synthesis approach to support ongoing exploration of scientific reasoning.  We have open-sourced the code and data for WildSci to enable scalable
and sustainable research in scientific reasoning.

\begin{table}[t]
\centering
\caption{Comparison of datasets by source type, domain coverage, size, question length, and originality. ‘New?’ indicates whether questions are newly generated or parsed from existing corpora.}
\resizebox{\linewidth}{!}{%
\begin{tabular}{lllrcc}
\toprule
\textbf{Dataset} & \textbf{Source} & \textbf{Domains} & \textbf{\# Q} & \textbf{Avg. Len.} & \textbf{New?} \\
\midrule
Camel-AI Science & GPT-4 (self-generated) & Phys, Bio, Chem & 60K & 30 $\pm$ 15 & Yes \\
Sci-Instruct     & Textbooks, problem sets, websites & Phys, Chem, Math, Formal Proofs & 254K & 41 $\pm$ 33 & No \\
SCP-116K         & Educational materials & Phys, Chem, Bio & 116K & 62 $\pm$ 74 & No \\
Natural Reasoning & Pretraining corpora & Multiple & 2.8M & 55 $\pm$ 21 & Yes \\
WildSci          & Peer-reviewed papers & Multiple (research focused) & 56K & 82 $\pm$ 19 & Yes \\
\bottomrule
\end{tabular}%
}

\label{tab:scienceqa_comparison}
\end{table}

%% file: ch_method.tex
\section{WildSci}

\subsection{Data Creation}
\begin{figure}
  \centering
  \includegraphics[width=\linewidth]{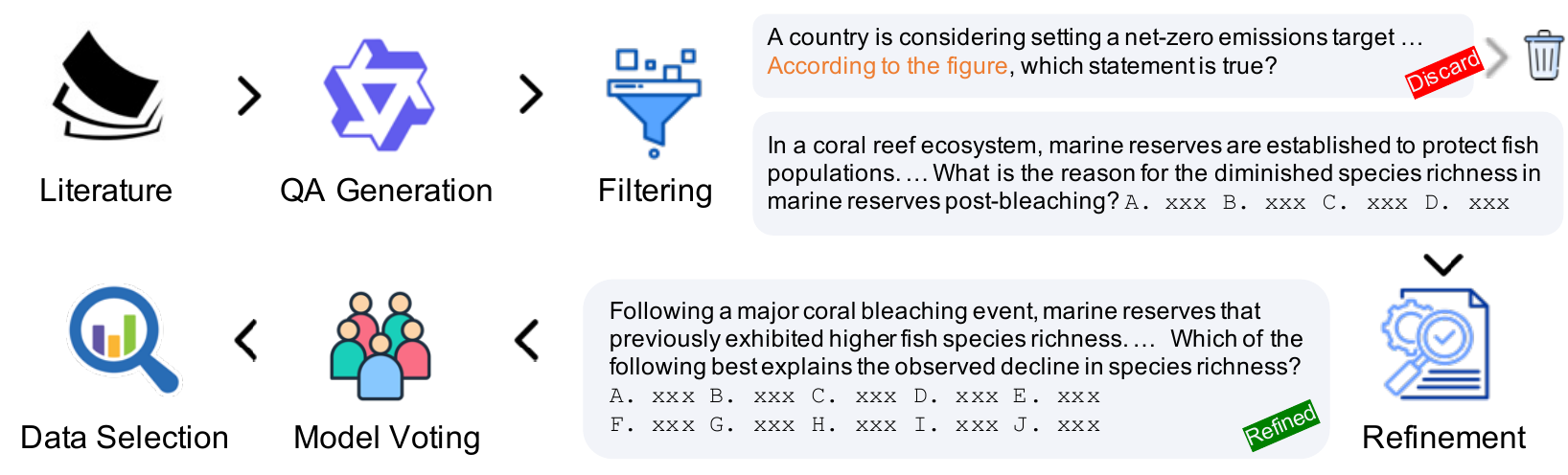}
  \caption{Overview of the data creation pipeline. Filtering is based on heuristic rules, while refinement expands the option space and rephrases questions to increase  diversity.}
  \label{fig:pipeline}
\end{figure}

An overview of our data creation pipeline is illustrated in Figure~\ref{fig:pipeline}. The entire process is fully automated using large language models, with multiple stages of filtering and refinement to ensure high-quality outputs. This automation enables our pipeline to generalize seamlessly to other scientific domains with accessible research literature.

\paragraph{Peer-reviewed Papers} 
We use publicly available, open-access articles from Nature Communications\footnote{\url{https://www.nature.com/ncomms/}} as the data source~\citep{DBLP:journals/corr/abs-2407-04903}.
The journal categorizes its content into five major areas and 72 subdomains. We reorganize these into nine broader disciplines following the taxonomy of SuperGPQA~\citep{DBLP:journals/corr/abs-2502-14739}. To ensure balanced coverage, we randomly sample a subset of papers from each category and generate three questions per paper based on its content. While the articles include both text and visuals, we focus exclusively on textual reasoning by using only the title, abstract, and main body, excluding all figures and tables.

\paragraph{QA Generation}

To enable a fully automated data synthesis pipeline, we employ a large language model to generate multiple-choice questions, corresponding answers and rationales directly from the paper content. The model is prompted to go through the full paper and create questions that reflect in-depth reasoning and understanding. Specifically, we instruct the model to produce context-independent questions. These are questions that can be answered without relying on figures, tables, or precise numerical details from the paper. This constraint ensures that the resulting questions emphasize knowledge-based and reasoning-intensive skills within a standalone context. The full prompt used for QA generation is provided in Appendix~\ref{app:prompts}.

\paragraph{Filtering}

To eliminate questions that require fine-grained recall or external references, we apply a set of heuristic and keyword-based filters. These filters remove items involving specific sections, detailed experimental results, or figure and table references (Figure~\ref{fig:pipeline}). 
Following previous work~\citep{DBLP:journals/corr/abs-2502-13124, eval-harness}, we also apply 13-gram deduplication against GPQA, SuperGPQA, and MMLU-Pro to eliminate semantically similar questions. 
The deduplication overlap rate is 0.0\%, suggesting that our dataset is both new and free of substantial overlap with existing resources.

\paragraph{Refinement}
The refinement stage focuses on enhancing question quality by increasing both their difficulty and diversity. Each question, along with its answer choices and rationale, is passed to LLM, which is prompted to paraphrase the question, eliminate surface-level cues, and expand the number of options (e.g., from 4 to 10 choices). This augmentation not only makes the questions more challenging but also reduces the chance of correct guesses during training. To better assess domain expert level reasoning, we explicitly instruct the model to remove well-known axioms or clues that are too obvious within a specific scientific field.

\paragraph{Model Voting}

Dataset quality can be negatively impacted by questions that are poorly constructed, lack necessary information, or are inherently unanswerable~\citep{DBLP:journals/corr/abs-2502-03461}. A practical way to assess answerability is to have models attempt to solve the questions. 
To validate the clarity of generated data, we apply model-based voting using an ensemble of open-source LLMs. Each model receives the question and answer choices, along with an additional fallback option: “None of the above / The question is unanswerable.” This allows models to flag ambiguous or ill-formed questions while trying to approach the solutions. Based on the model voting results, we discard questions where the majority of models select the unanswerable option.

\paragraph{Data Selection}
We further categorize the questions by the level of agreement among ensemble models, using this as a proxy for clarity and difficulty:
This grouping supports more controlled training setups, enabling models to learn from data with varying complexity:\\
\textbf{All Aligned} All model responses match the synthetic answer, indicating the question is clear and easily answerable.\\
\textbf{Majority Aligned} The majority vote of the model responses matches the original labels. These questions are valid but more challenging, as models do not consistently arrive at the correct answer.\\
\textbf{Majority Divergent} 
The majority of responses differ from the original labels. This indicates either that the questions are  challenging or that the synthetic labels may be incorrect, with the majority answers potentially representing more reasonable solutions.\\
\textbf{All Divergent} No single answer is selected by more than half of the ensemble responses. Such questions are likely highly difficult or ambiguous, resulting in diverse interpretations.

\subsection{Statistics}


\begin{minipage}[t]{0.58\textwidth}
  \hspace*{-0.4cm} \includegraphics[width=\linewidth]{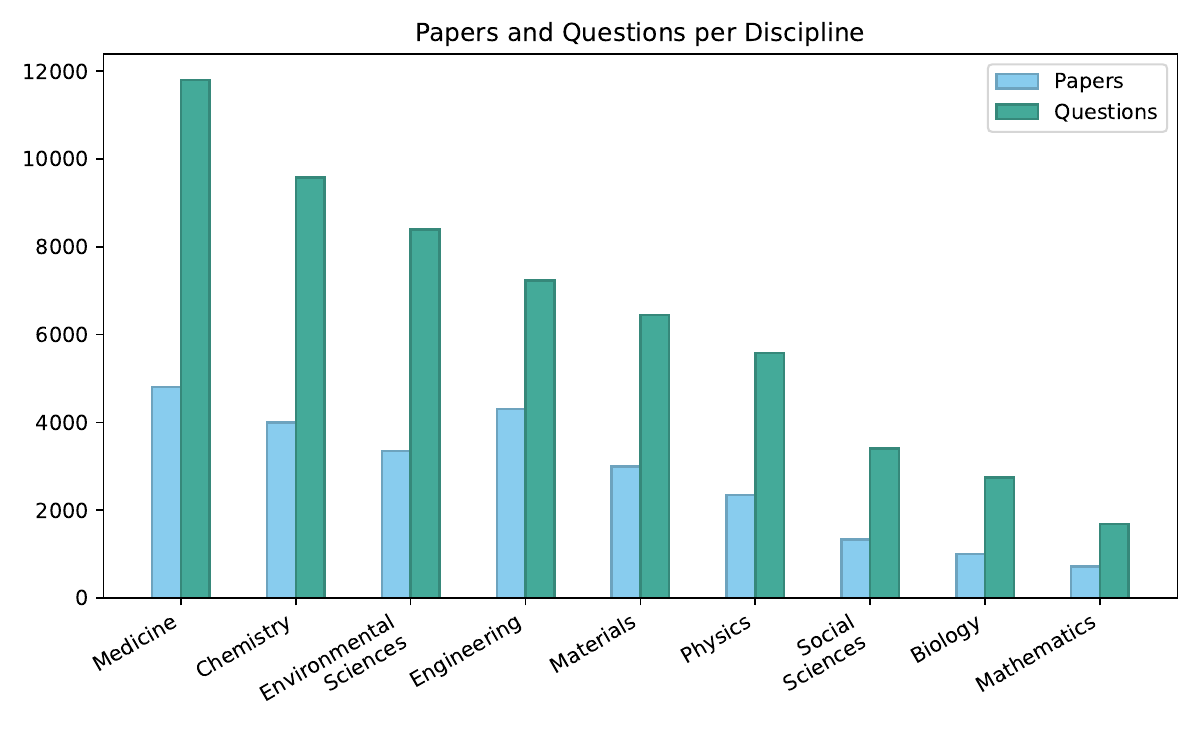}
  \captionof{figure}{Comparison of the number of papers and gen- \\ erated questions across different disciplines.}
  \label{fig:paper_q}
\end{minipage}\hfill
\begin{minipage}[t]{0.42\textwidth}
    \includegraphics[width=\linewidth]{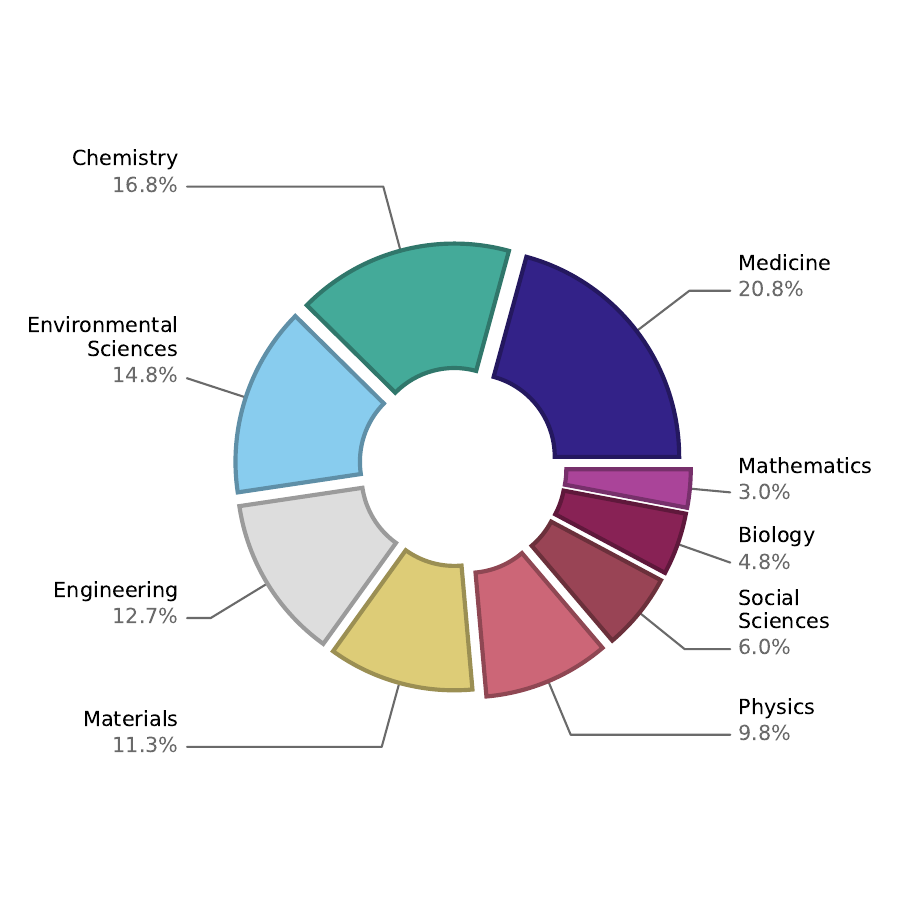}  
    \captionof{figure}{Distribution of questions after filtering and refinement in WildSci.}
    \label{fig:q_dist}
\end{minipage}

\paragraph{Domains}

The distribution of selected papers and the resulting question dataset after filtering and refinement are illustrated in Figure~\ref{fig:paper_q} and Figure~\ref{fig:q_dist}. Note that many subdomains are interdisciplinary, for example, Biochemistry can fall under both Biology and Chemistry. This classification serves primarily to provide a high-level overview of the dataset composition. A comprehensive taxonomy and detailed subdomain statistics are included in Appendix~\ref{app:stat_detail}.

\paragraph{Length}

We measure question length by counting the number of words in each question, excluding the  options. While word count does not directly indicate difficulty, it serves as a proxy for the complexity and richness of the question descriptions. 
Grounded in real-world research, questions in WildSci often include background information to provide context. Compared to other datasets in Table~\ref{tab:scienceqa_comparison}, WildSci has the longest questions, with a mean length of 81.74 words. Among different disciplines, Mathematics has the highest average word count (93.9), followed by Physics (89.8) and Engineering (88.6). In contrast, questions in Biology are the shortest on average, with 69.9 words. A detailed distribution of question lengths is provided in Appendix~\ref{app:stat_detail}.

\subsection{Training}

\paragraph{Reward Design}
In order to enable RLVR in science domains, WildSci allows obtaining verifiable rewards during training through simple option answer matching. 
We denote the original labels produced by our data creation pipeline as synthetic labels, denoted $y_\texttt{syn}$. For each question, there is exactly one such label generated automatically from the pipeline. 
In this setting, we define the reward function based solely on matching the prediction $\hat y$ and the synthetic label $y_{\texttt{syn}}$:

\begin{equation}
\mathcal{R}_{\texttt{syn}}(\hat y) = 
\begin{cases}
1.0, & \text{if } \hat y = y_{\texttt{syn}}, \\
0.0, & \text{otherwise}.
\end{cases}
\end{equation}\label{eq:reward}

To prevent models from memorizing option positions, we randomly shuffle the choices in each training epoch, ensuring a balanced distribution and reducing overfitting to option labels.

\paragraph{Training Algorithm}

Following previous work that advances mathematical reasoning of LLMs, we adopt Group Relative Policy Optimization (GRPO) as the training algorithm~\citep{DBLP:journals/corr/abs-2402-03300}.

%% file: ch_experiments.tex
\section{Experiments}

\subsection{Evaluation Benchmarks}

To assess the scientific reasoning capabilities of language models, we evaluate them on a suite of benchmarks targeting scientific question answering. We report accuracy for each dataset by checking whether the final selected option in the model's response is correct.

\paragraph{WildSci-Val} We construct an in-domain validation set using All Aligned questions from WildSci. Specifically, we randomly sample 100 questions from each of the 9 disciplines, resulting in a total of 900 questions. Each question includes 10 available options, with exactly one correct answer.

\paragraph{GPQA} 

To mitigate potential bias introduced by answer choice order, we introduce \textbf{GPQA-Aug}, an augmented variant of the original GPQA-Diamond dataset~\citep{DBLP:journals/corr/abs-2311-12022}. For each of the 198 questions, we generate four versions by permuting the answer options so that the correct answer appears once in each of the four positions, resulting in a total of 792 examples.

\paragraph{SuperGPQA} SuperGPQA~\citep{DBLP:journals/corr/abs-2502-14739} is a large-scale benchmark designed to evaluate knowledge and reasoning across 285 graduate-level disciplines, with 26,529 questions in total.

\paragraph{MMLU-Pro} MMLU-Pro~\citep{DBLP:conf/nips/WangMZNCGRAHJLK24} is a dataset of 12,032 questions that builds upon the original MMLU~\citep{DBLP:conf/iclr/HendrycksBBZMSS21} by shifting its focus to reasoning-based questions.

\subsection{Data Creation}
We adopt \texttt{Qwen2.5-32B-Instruct}~\citep{DBLP:journals/corr/abs-2412-15115} and \texttt{Qwen3-32B}~\citep{DBLP:journals/corr/abs-2402-03300} as the default model in our data generation pipeline, balancing quality and cost-effectiveness, as supported by findings in \citet{Moshkov2025AIMO2WS}. For the model voting stage, we additionally employ \texttt{Mistral-Small-24B-Instruct-2501}, selected for their similar performance in scientific reasoning tasks. We generate 4 responses per model using temperature of 0.8, resulting in a total of 8 responses used in the voting process.

%% file: ch_results.tex
\section{Results and Analysis}

\subsection{Improving Science Reasoning Abilities}

\begin{table}[ht]
\centering
\caption{Performance comparison across different benchmarks. The Average column is computed across the three public benchmarks GPQA-Aug, SuperGPQA and MMLU-Pro. 
Maj. Aligned stands for the Majority Aligned subset of WildSci.
}
\resizebox{\linewidth}{!}{%
\begin{tabular}{lccccc}
\toprule
\textbf{Model} & \textbf{WildSci-Val} & \textbf{GPQA-Aug} & \textbf{SuperGPQA} & \textbf{MMLU-Pro} & \textbf{Average} \\
\midrule
Qwen2.5-1.5B-Instruct & 46.70\textsubscript{1.98} & 23.98\textsubscript{0.83} & 18.10\textsubscript{0.63} & 31.47\textsubscript{0.84} & 24.52\textsubscript{0.76} \\
+ WildSci All Aligned & \textbf{80.48}\textsubscript{0.26} & \textbf{28.95}\textsubscript{0.08} & 23.85\textsubscript{0.25} & \textbf{42.54}\textsubscript{0.28} & \textbf{31.78}\textsubscript{0.20} \\
+ WildSci Maj. Aligned & 80.33\textsubscript{0.11} & 25.76\textsubscript{0.44} & \textbf{24.41}\textsubscript{0.06} & 41.95\textsubscript{0.03} & 30.71\textsubscript{0.17} \\
\midrule
Qwen2.5-3B-Instruct & 72.45\textsubscript{0.40} & 28.03\textsubscript{1.97} & 23.21\textsubscript{0.08} & 44.18\textsubscript{0.15} & 31.80\textsubscript{0.73} \\
+ WildSci All Aligned & \textbf{85.00}\textsubscript{0.40} & \textbf{33.04}\textsubscript{0.86} & \textbf{26.39}\textsubscript{0.38} & \textbf{49.33}\textsubscript{0.27} & \textbf{36.25}\textsubscript{0.50} \\
+ WildSci Maj. Aligned & 84.71\textsubscript{0.06} & 30.98\textsubscript{0.89} & 26.01\textsubscript{0.24} & 48.66\textsubscript{0.10} & 35.22\textsubscript{0.41} \\
\bottomrule
\end{tabular}
}
\label{tab:main}
\end{table}

In this section, we apply GRPO to train models on different subsets of WildSci.
Table~\ref{tab:main} reports the main results across multiple evaluation sets. WildSci-Val shows in-domain performance, while the Average column reflects mean accuracy across three out-of-domain science benchmarks: GPQA-Aug, SuperGPQA, and MMLU-Pro.
Training \texttt{Qwen2.5-1.5B-Instruct} on the All Aligned subset significantly boosts in-domain accuracy from 46.7\% to 80.48\%. It also improves generalization to public benchmarks, with gains of 4.97\%, 5.75\%, and 11.07\% on GPQA-Aug, SuperGPQA, and MMLU-Pro respectively, resulting in an average increase of 7.26\%.
Consistent gains are also observed with 3B models, where WildSci yields an average improvement of 4.45\% across science benchmarks. While training on the Majority Aligned subset also brings similar improvements, its performance is slightly lower than the All Aligned subset. Overall, we do not observe substantial performance differences across subsets, aligning with findings from \citet{Wang2025ReinforcementLF}. A detailed distribution and comparison of subsets is provided in Figure~\ref{fig:q_type} and Appendix~\ref{app:data_splits}.

\begin{wrapfigure}{r}{0.45\textwidth}  
  \centering
  \includegraphics[width=0.42\textwidth]{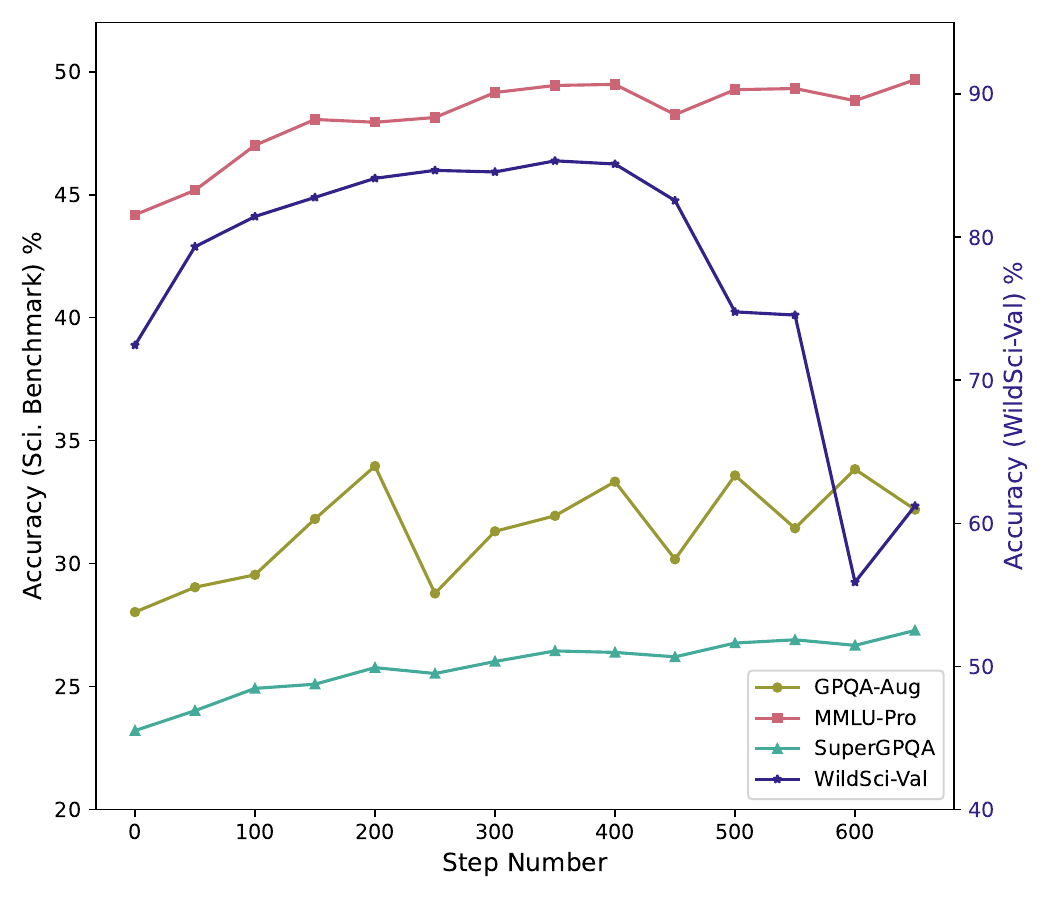}
  \caption{Performance trends on validation and test sets during training of the 3B model on WildSci All Aligned. The model exhibits continued generalization on test sets even after overfitting on the validation set.}
  \label{fig:ood3b}
  \vspace{-20pt}
\end{wrapfigure}

Furthermore, we observe that \textbf{models continue to generalize even after overfitting on the validation set}.
As illustrated in Figure~\ref{fig:ood3b}, while the model's performance on in-domain validation set begins to decline sharply after step 400, its accuracy on OOD test datasets continues to improve. This pattern mirrors the post-saturation generalization phenomenon identified by \citet{Wang2025ReinforcementLF} in math domain. The presence of this generalization behavior indicates WildSci's potential as an ideal testbed for investigating RL reasoning in scientific contexts.

\subsection{Domain-Specific Performance Dynamics}

\begin{figure}[t]
  \centering
  \includegraphics[width=\linewidth]{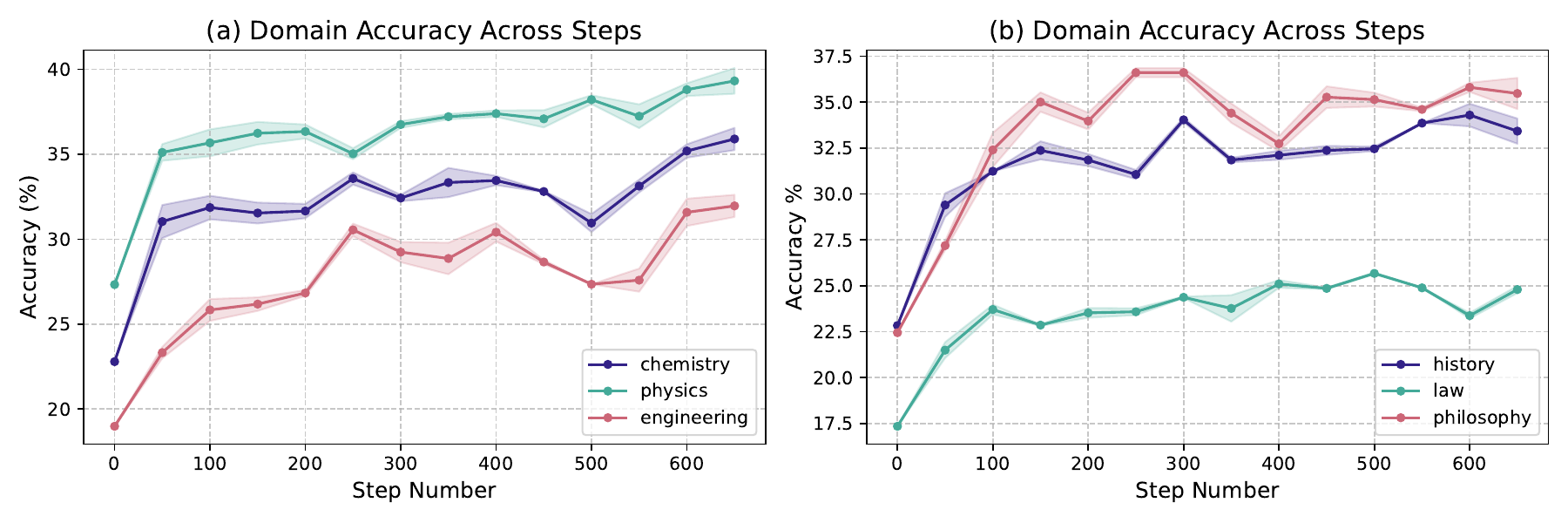}
  \caption{Domain-specific accuracy trends on MMLU-Pro during training. We report mean accuracy across three runs for each domain. Shaded regions indicate standard deviation. (a) shows steady improvements in domains with higher WildSci coverage (chemistry, physics, engineering), while (b) illustrates more variable performance in domains with lower coverage (law, history, philosophy).}
  \label{fig:per_domain}
\end{figure}

Figure~\ref{fig:per_domain} illustrates the performance trends across subdomains of MMLU-Pro during training on a 1.5B model with WildSci Majority Aligned data. We observe a sharp increase in accuracy within the first 100 steps, likely due to the model's rapid adaptation to the multiple-choice answer format. After this initial phase, performance stabilizes but exhibits domain-specific dynamics.
Interestingly, we find that in domains such as chemistry, physics, and engineering, where WildSci provides more coverage, accuracy continues to improve steadily throughout training (Figure~\ref{fig:per_domain}(a)). In contrast, domains with sparser data coverage, such as law, history, and philosophy, show more fluctuation and slower gains (Figure~\ref{fig:per_domain}(b)).
The observed performance differences highlight the need for more balanced data coverage across domains to ensure uniform model improvements. A detailed breakdown of domain-level performance on SuperGPQA and MMLU-Pro can be found in Appendix~\ref{app:per_domain}.

\subsection{Analyzing Format Alignment and Reasoning Improvement}

As our training data consists of multiple-choice questions, the model naturally adapts to the expected answer format and becomes familiar with the task structure. This raises the question: \textit{are the observed improvements on science reasoning benchmarks only a result of format alignment, or do they reflect gains in reasoning ability?}

To disentangle the effect of format adaptation from actual reasoning improvement, we track the model's adherence to the expected answer format over training steps. Figure~\ref{fig:res_format} presents both the proportion of responses from which a valid final answer can be reliably extracted and the corresponding accuracy trends.

Our analysis shows that even by step 5, the model achieves a high extraction success rate of 88.86\%, which quickly converges to around 95\% within just 20 steps. This rapid convergence suggests that \textbf{the model learns the structural template of multiple-choice answers very early during training}.
Furthermore, we continue to observe steady improvements in accuracy beyond this point, indicating that the later-stage performance gains cannot be attributed solely to improved format adaptation. Instead, these results indicate \textbf{a gradual enhancement in the model’s reasoning capabilities} as it attempts and learns from more diverse responses during training.


\begin{minipage}[t]{0.4\textwidth}
    \includegraphics[width=\linewidth]{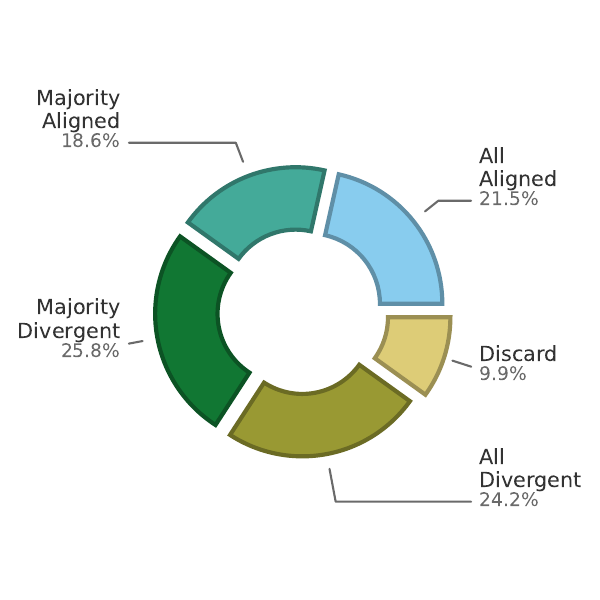}  
    \captionof{figure}{Proportion of different \\data splits within WildSci.}
    \label{fig:q_type}
\end{minipage}\hfill
\begin{minipage}[t]{0.6\textwidth}
  \includegraphics[width=\linewidth]{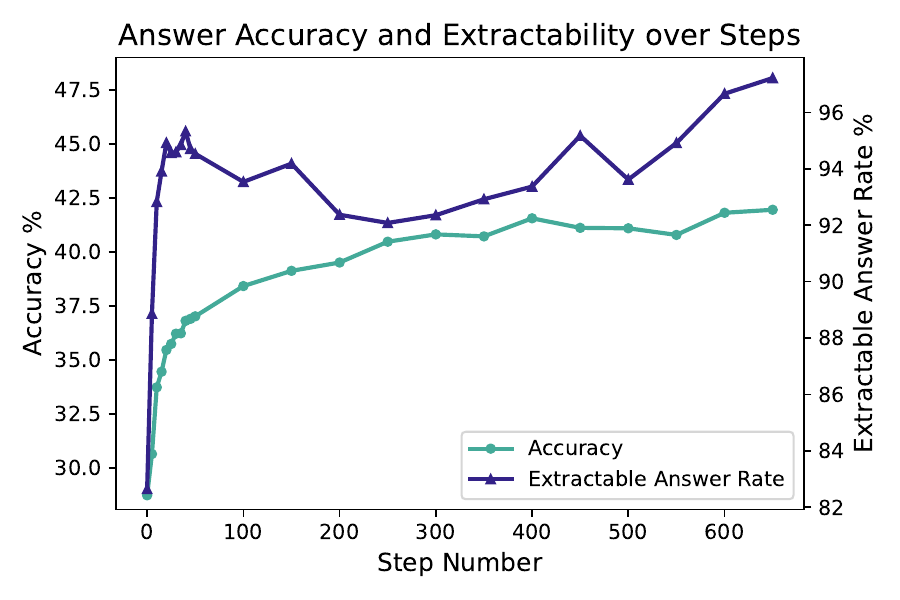}
  \captionof{figure}{Answer accuracy and extractability measured across training steps on MMLU-Pro.}
  \label{fig:res_format}
\end{minipage}

\subsection{Ablation Analysis}

In this section, we conduct an ablation analysis to examine the impact of question refinement on final model performance. During the refinement phase, we enhance both the semantic diversity of questions and the breadth of the answer option space. The results are presented in Table~\ref{tab:no_refine}. In both settings, we train the \texttt{Qwen2.5-1.5B-Instruct} model using All Aligned questions from WildSci.

Even without refinement, the model achieves an improved average performance of 28.83\%, representing a 4.31\% gain over the baseline average accuracy of 24.52\%. However, with only four options per question in the training set, the model only achieves 66.63\% on the valid set which contains 10 options per question. Notably, the expanded answer space introduced during refinement not only balances the distribution among the answer space, but also reduces the likelihood of the model arriving at the correct answer through random guessing, thereby encouraging more robust learning and leading to further performance improvements on OOD science benchmarks.

\begin{table}[ht]
\centering
\caption{Performance comparison before and after refinement stage. The average is computed across the three public benchmarks GPQA-Aug, SuperGPQA and MMLU-Pro. We use All Aligned data as the training data in both settings.}
\resizebox{\linewidth}{!}{%
\begin{tabular}{lccccc}
\toprule
\textbf{Setting} & \textbf{WildSci-Val} & \textbf{GPQA-Aug} & \textbf{SuperGPQA} & \textbf{MMLU-Pro} & \textbf{Average} \\
\midrule
WildSci & \textbf{80.48}\textsubscript{0.26} & \textbf{28.95}\textsubscript{0.08} & \textbf{23.85}\textsubscript{0.25} & \textbf{42.54}\textsubscript{0.28} & \textbf{31.78}\textsubscript{0.20} \\
WildSci w/o refinement & 66.63\textsubscript{0.06} & 27.94\textsubscript{0.32} & 20.44\textsubscript{0.04} & 38.11\textsubscript{0.37} & 28.83\textsubscript{0.24} \\
\bottomrule
\end{tabular}
}
\label{tab:no_refine}
\end{table}

\subsection{Mixed Training with MATH}

To assess the transferability between mathematical and scientific reasoning, we conduct experiments using the MATH dataset~\citep{DBLP:conf/nips/HendrycksBKABTS21} for both training and evaluation of math reasoning abilities. As shown in Table~\ref{tab:mix_math}, model trained solely on MATH achieves improved performance on math-specific tasks but fails to generalize to science reasoning benchmarks. In contrast, training solely on WildSci also leads to improvements in math reasoning, indicating its broader effectiveness.
We then perform mixed training using both MATH and WildSci, denoted as ``MATH+WildSci”. The resulting model demonstrates improved performance on science benchmarks while preserving its math reasoning ability. Moreover, it also improves accuracy on GPQA-Aug, highlighting the effect of combining science and math reasoning data. \textbf{These results suggest that WildSci complements existing reasoning datasets and helps enhance model generalization across diverse reasoning domains}.

\begin{table}[ht]
\centering
\caption{Performance comparison when training 1.5B model with MATH. The average is computed across the three public benchmarks GPQA-Aug, SuperGPQA and MMLU-Pro. We use Maj. Aligned subset as the training data for WildSci.}
\resizebox{\linewidth}{!}{%
\begin{tabular}{lcccccc}
\toprule
\textbf{Setting} & \textbf{WildSci-Val} & \textbf{GPQA-Aug} & \textbf{SuperGPQA} & \textbf{MMLU-Pro} & \textbf{Sci. Avg.} & \textbf{MATH} \\
\midrule
Qwen2.5-1.5B-Instruct & 46.70\textsubscript{1.98} & 23.98\textsubscript{0.83} & 18.10\textsubscript{0.63} & 31.47\textsubscript{0.84} & 24.52\textsubscript{0.76} & 55.47\textsubscript{0.83} \\
WildSci & 80.33\textsubscript{0.11} & 25.76\textsubscript{0.44} & 24.41\textsubscript{0.06} & 41.95\textsubscript{0.03} & 30.71\textsubscript{0.17} & 57.00\textsubscript{0.60}\\
MATH & 36.96\textsubscript{1.18} & 20.54\textsubscript{0.44} & 17.47\textsubscript{0.05} & 28.86\textsubscript{0.05} & 22.29\textsubscript{0.18} & 58.92\textsubscript{0.87} \\
\rowcolor{gray!30} MATH + WildSci & 78.41\textsubscript{0.34} & 29.76\textsubscript{0.45} & 24.25\textsubscript{0.12} & 42.36\textsubscript{0.09} & 32.12\textsubscript{0.22} &  58.73\textsubscript{0.99}\\
\bottomrule
\end{tabular}
}
\label{tab:mix_math}
\end{table}

\subsection{Scientific Reasoning Type Analysis}

To characterize the types of reasoning required in WildSci, two authors manually reviewed 200 randomly sampled questions and developed a classification scheme. Each question was categorized into one of the following three scientific reasoning types:
(1) Mathematical calculations and derivations: Questions that require performing numerical computations or symbolic derivations.
(2) Model design, method analysis, or conceptual understanding: Questions that involve understanding scientific models, evaluating methodologies, or grasping abstract concepts.
(3) Causal reasoning and mechanism inference: Questions that ask about the underlying causes, mechanisms, or consequences of a phenomenon.

To extend this categorization to the full dataset, we use \texttt{Qwen2.5-32B-Instruct} to automatically classify all questions in WildSci. The resulting distribution is as follows: 40.00\% of the questions involve numerical calculation, 37.59\% require causal inference, and 22.41\% focus on model analysis or conceptual understanding.
While mathematical reasoning is common in other datasets, WildSci emphasizes higher-order scientific reasoning skills such as causal inference and model-based analysis. These types of abilities are essential in research-oriented problem solving, reflecting the complexity and diversity of scientific reasoning in our dataset.

%% file: ch_conclusion.tex
\section{Limitations}\label{sec:limitation}

While involving domain-specific knowledge in the questions, some numerical questions in our dataset are relatively simple. Although complex mathematical reasoning is not the primary focus of this work, combining scientific knowledge with deeper quantitative reasoning is a promising future direction.
Another limitation lies in the use of a multiple-choice question (MCQ) format. Despite our efforts to carefully design questions and diversify answer choices to reduce superficial patterns, the format inherently introduces the risk of models exploiting spurious heuristics. Evaluating and verifying open-ended research questions such as causal reasoning and analysis remains an open challenge in advancing scientific reasoning abilities of LLMs.

\section{Conclusion}
We present WildSci, a dataset of verifiable scientific reasoning questions automatically generated from peer-reviewed papers. Our pipeline synthesizes questions across 9 disciplines and 26 subdomains, using model voting for quality control and pairing each question with 10 answer options. RLVR training on WildSci leads to improved performance on multiple science reasoning benchmarks. Training dynamics further confirm the effectiveness and generalizability of our data. As scientific literature continues to grow, WildSci offers a sustainable approach to converting real world research articles into valuable data for advancing scientific reasoning in language models.

%% file: ch_appendix.tex
\newpage
\section{Experiment Settings}\label{sec:exp}

We train our model using the GRPO algorithm implemented in the verl~\citep{sheng2024hybridflow} framework. We set a maximum response length of up to 8192 tokens and applying left-side truncation. During rollout, we sample 8 responses per prompt with a temperature of 1.0. The model is trained with a learning rate of $5 \times 10^{-7}$ and is updated with AdamW optimizer~\citep{Loshchilov2017DecoupledWD}. We mostly adopt the training settings from \citet{DBLP:journals/corr/abs-2503-18892} and do not perform additional hyperparameter tuning. 
All experiments are conducted on a server equipped with 8 NVIDIA A100 GPUs, each with 40GB of memory. Training the 1.5B model for 700 steps takes one day using 4 GPUs, while the 3B model requires two days on all 8 GPUs for 700 steps. We select checkpoints for evaluation based on the in-domain validation set WildSci-Val and the final training checkpoint at step 700.
During the evaluation phase, we set the temperature to 0.0 using the vLLM library to eliminate randomness in model predictions~\citep{kwon2023efficient}.
To show potential variance in inference, we report the mean and standard deviation over three independent runs.

\section{Related Work}

\subsection{Science Reasoning Data}

Domain-specific datasets have significantly advanced model reasoning, especially in mathematics~\citep{DBLP:conf/iclr/YuJSYLZKLWL24, DBLP:conf/iclr/YueQZFH00C24, DBLP:conf/iclr/ToshniwalDMKAG25, DBLP:journals/corr/abs-2502-17387, Zhou2025MegaMathPT}. Recent work has extended coverage to core natural sciences like physics, chemistry, and biology~\citep{DBLP:conf/nips/LiHIKG23, DBLP:journals/corr/abs-2501-15587}. SciInstruct~\citep{DBLP:conf/nips/ZhangHZDYWYD024} collects questions from textbooks, exams, and problem sets, while \citet{DBLP:conf/nips/YueZZC24, DBLP:journals/corr/abs-2502-13124, generalreasoner} synthesize questions from web text and pretraining corpora. In contrast, WildSci constructs questions from peer-reviewed research papers, targeting in-the-wild scientific problems and enabling generalization to long-tail domains through an automatic data synthesis pipeline.

\subsection{Reinforcement Learning in Reasoning}

Reinforcement learning has advanced reasoning in areas like mathematics and coding, where simple rule-based rewards have proven effective~\citep{ DBLP:journals/corr/abs-2501-12599, DBLP:journals/corr/abs-2503-24290, DBLP:journals/corr/abs-2503-14476, DBLP:journals/corr/abs-2503-20783, DBLP:journals/corr/abs-2503-18892, DBLP:journals/corr/abs-2402-03300, DBLP:conf/emnlp/LiuGHZQZ22}. In science domains, open-ended questions can make rule-based verification challenging. For efficiency and simplicity, we reformat open science research questions as MCQs. While concurrent work validates the use of MCQs for RL training~\citep{Akter2025NemotronCrossThinkSS}, WildSci specifically focuses on grounding its questions within real-world scientific literature. This approach not only ensures domain specificity but also provides a convenient training signal and facilitates generalization across diverse domains.

\subsection{Science Reasoning Evaluation}

Recent studies have developed benchmarks to address the limited evaluation of LLMs in the science domain~\citep{DBLP:journals/corr/abs-2502-14739,DBLP:conf/emnlp/Fei0ZZHHZ0YSG024, DBLP:conf/chil/PalUS22,jin2020disease,DBLP:journals/corr/abs-2412-15194,DBLP:journals/corr/abs-2411-07240,DBLP:conf/iclr/Gao0YCMDLMCXTWZ25,putnam_axiom2025}. SciBench~\citep{DBLP:conf/icml/WangHL0ZSLZS024} provides open-ended, free-response collegiate-level problems that require multi-step reasoning abilities, including advanced mathematical derivation and understanding of scientific concepts, in chemistry, physics, and math. Other benchmarks such as SciKnowEval~\citep{DBLP:journals/corr/abs-2406-09098} and SciAccess~\citep{DBLP:journals/corr/abs-2403-01976} evaluate LLMs in memorization, comprehension and reasoning across various science domains. Domain-specialized datasets like MaScQA~\citep{DBLP:journals/corr/abs-2308-09115} and UGPhysics~\citep{DBLP:journals/corr/abs-2502-00334} further evaluate models within a specific subject. ResearchBench~\citep{DBLP:journals/corr/abs-2503-21248} provides a benchmark to test LLMs on generating innovative scientific hypotheses, introducing a framework to assess inspiration retrieval, hypothesis composition, and hypothesis ranking. Although these are effective metrics to evaluate scientific proficiency, only a few domains are actively involved, mainly chemistry, biology, materials science, and physics. WildSci encompasses a wide range of scientific disciplines to support a comprehensive dataset. Through a multidisciplinary and multiple-choice approach, we envision WildSci to be an impactful and scalable resource for scientific research and model development.

\section{Data Quality Analysis}

\subsection{Validation of Synthetic Annotations}

To evaluate the reliability of the synthetic labels in WildSci, we conduct a validation study using two strong commercial models, Gemini-2.5-Pro and Gemini-2.5-Flash~\citep{DBLP:journals/corr/abs-2507-06261}. We randomly sample 500 questions from both the All-Aligned and Majority-Aligned subsets and independently prompt each model to generate answers. We then compare their responses against our synthetic labels to measure answer agreement.

On the All-Aligned subset, Flash and Pro agree with our synthetic labels in 95.0\% and 96.0\% of cases, respectively. The two Gemini models agree with each other in 94.0\% of cases, and among these agreements, 98.9\% match our synthetic labels. 
On the Majority-Aligned subset, Flash and Pro achieve 78.6\% and 80.2\% agreement with our labels, while their mutual agreement rate is 80.6\%, with 88.8\  of those matches aligning with our labels.

These results demonstrate that \textbf{WildSci's filtering and model-voting procedure produces annotations that are highly consistent with those from substantially stronger models}. Combined with the observed performance gains when training open source models, this analysis supports that the synthetic data quality in WildSci is sufficiently high for effective model training.

\subsection{Question Redundancy}

We assessed question similarity using SentenceTransformer~\citep{reimers-2019-sentence-bert} and computed cosine similarity between question pairs within the same domain. We only consider question pairs with similarity $\geq$ 0.9, and we found that highly similar pairs account for 2.7\% in “All-aligned” and 2.3\% in “Majority-aligned”, indicating low redundancy. While these pairs often share surface-level phrasing (e.g., from the same paper), manual inspection shows they frequently assess distinct concepts. For example, the following pair has a similarity score of 0.902 but asks fundamentally different things.

To assess potential redundancy in the WildSci dataset, we measured pairwise question similarity within each domain using the SentenceTransformer model and computed cosine similarity between question embeddings. We considered question pairs with similarity scores $\geq 0.9$ as highly similar.
Such pairs constitute only 2.7\% of the All-Aligned subset and 2.3\% of the Majority-Aligned subset, indicating low redundancy across questions. Manual inspection further reveals that these high-similarity pairs often share surface-level phrasing (e.g., derived from the same paper section) yet still assess distinct scientific concepts or reasoning steps.

To illustrate this, Table~\ref{tab:example_similarity} presents a representative example of two highly similar questions (cosine similarity = 0.902). Despite their surface resemblance, the two queries probe distinct reasoning objectives—quantitative estimation versus selectivity comparison—highlighting that lexical overlap does not necessarily imply conceptual redundancy.

\begin{table*}[h]
\centering
\caption{Example of a high-similarity question pair (cosine similarity = 0.902) that differs in reasoning objective.}
\label{tab:example_similarity}
\begin{tabular}{lp{0.85\linewidth}}
\toprule
\textbf{Q1} & A hybrid absorption–adsorption system is being evaluated for CO\textsubscript{2} capture using a slurry composed of ZIF-8 suspended in a 2-methylimidazole-based glycol solution. \textit{What is the total amount of CO\textsubscript{2} (in moles) that can be captured by 1 liter of the slurry under ideal behavior assumptions?} \\
\addlinespace[3pt]
\textbf{Q2} & A hybrid absorption–adsorption system for CO\textsubscript{2} capture employs a slurry of ZIF-8 in a 2-methylimidazole-based glycol solution. \textit{What is the ratio of the amount of CO\textsubscript{2} captured to the amount of N\textsubscript{2} captured? Assume that the selectivity is defined as the ratio of the partition coefficients of the two gases.} \\
\bottomrule
\end{tabular}

\end{table*}

Overall, this analysis confirms that WildSci maintains \textbf{high question diversity} despite its large scale, with minimal duplication and strong coverage of distinct reasoning patterns even among lexically similar items.

\subsection{Validity and Difficulty}

To further evaluate data quality control, we conducted an additional analysis using Gemini-2.5-Pro~\citep{DBLP:journals/corr/abs-2507-06261} to assess the validity and difficulty of questions across the four WildSci subsets. For each subset, we randomly sampled 500 examples and prompted the model to independently rate question clarity and difficulty.

\paragraph{Validity Evaluation}
For validity, the model was instructed to classify each question as either \emph{good and clear} or \emph{unanswerable}. As summarized in Table~\ref{tab:validity}, 96.6\% of questions in the All-Aligned (A.A.) subset and 89.4\% in the Majority-Aligned (M.A.) subset were rated as good and clear, compared to 87.4\% and 71.2\% for the Majority-Divergent (M.D.) and All-Divergent (A.D.) subsets, respectively.

These results indicate that the vast majority of WildSci questions are coherent and answerable, particularly in the more reliable subsets. This finding aligns with the design intuition behind our model-voting phase and helps explain why the All-Aligned and Majority-Aligned subsets lead to stronger downstream training performance.

\begin{table*}[h]
\centering
\caption{Validity ratings across the four WildSci subsets. A large majority of questions, especially in the aligned subsets, were rated as clear and answerable.}
\label{tab:validity}
\setlength{\tabcolsep}{6pt}
\begin{tabular}{lcccc}
\toprule
\textbf{Subset} & \textbf{All-Aligned} & \textbf{Majority-Aligned} & \textbf{Majority-Divergent} & \textbf{All-Divergent} \\
\midrule
Good \& Clear (\%) & 96.6 & 89.4 & 87.4 & 71.2 \\
Unanswerable (\%)  & 3.4  & 10.6 & 12.6 & 28.8 \\
\bottomrule
\end{tabular}
\end{table*}

\paragraph{Difficulty Evaluation}
We further asked Gemini-2.5-Pro to rate each question’s difficulty on a five-level scale, from \textbf{Level 1 (Trivial)} to \textbf{Level 5 (Expert)}. As shown in Table~\ref{tab:difficulty}, a substantial portion of questions in the All-Aligned (40.8\%) and Majority-Aligned (59.0\%) subsets were classified as Level 4--5, requiring undergraduate- or graduate-level domain expertise.

Overall, this analysis demonstrates that the question filtering and model-voting stages effectively select for both clarity and appropriate difficulty. The All-Aligned and Majority-Aligned subsets, in particular, comprise questions that are not only valid and well-posed but also intellectually challenging, supporting effective model training in various science reasoning domains.

\begin{table}[h]
\centering
\caption{Distribution of difficulty levels across subsets. Levels 4–5 indicate field-specific undergraduate or graduate-level reasoning.}
\label{tab:difficulty}
\begin{tabular}{lccccc}
\toprule
\textbf{Subset} & \textbf{L1} & \textbf{L2} & \textbf{L3} & \textbf{L4} & \textbf{L5} \\
\midrule
A.A.        & 0.8 & 28.6 & 29.8 & 38.4 & 2.4 \\
M.A.   & 0.0 & 12.8 & 28.2 & 52.4 & 6.6 \\
M.D. & 0.2 & 11.8 & 27.6 & 55.6 & 4.8 \\
A.D.      & 0.0 & 6.6  & 22.8 & 59.8 & 10.8 \\
\bottomrule
\end{tabular}
\end{table}

\begin{table*}[h]
\centering
\caption{Difficulty rubric used in the difficulty evaluation.}
\label{tab:rubric}
\begin{tabular}{cl}
\toprule
\textbf{Level} & \textbf{Description} \\
\midrule
5 & Very Challenging: Expert-level (graduate/PhD) reasoning required. \\
4 & Difficult: Field-specific undergraduate-level understanding. \\
3 & Moderate: General background or non-specialist undergraduate-level knowledge. \\
2 & Easy: General science or high-school-level knowledge. \\
1 & Simple: Common-sense or minimal prior knowledge suffices. \\
\bottomrule
\end{tabular}
\end{table*}

\section{Statistics of Subdomains}\label{app:stat_detail}

\begin{figure}
  \centering
  \includegraphics[width=1.0\linewidth]{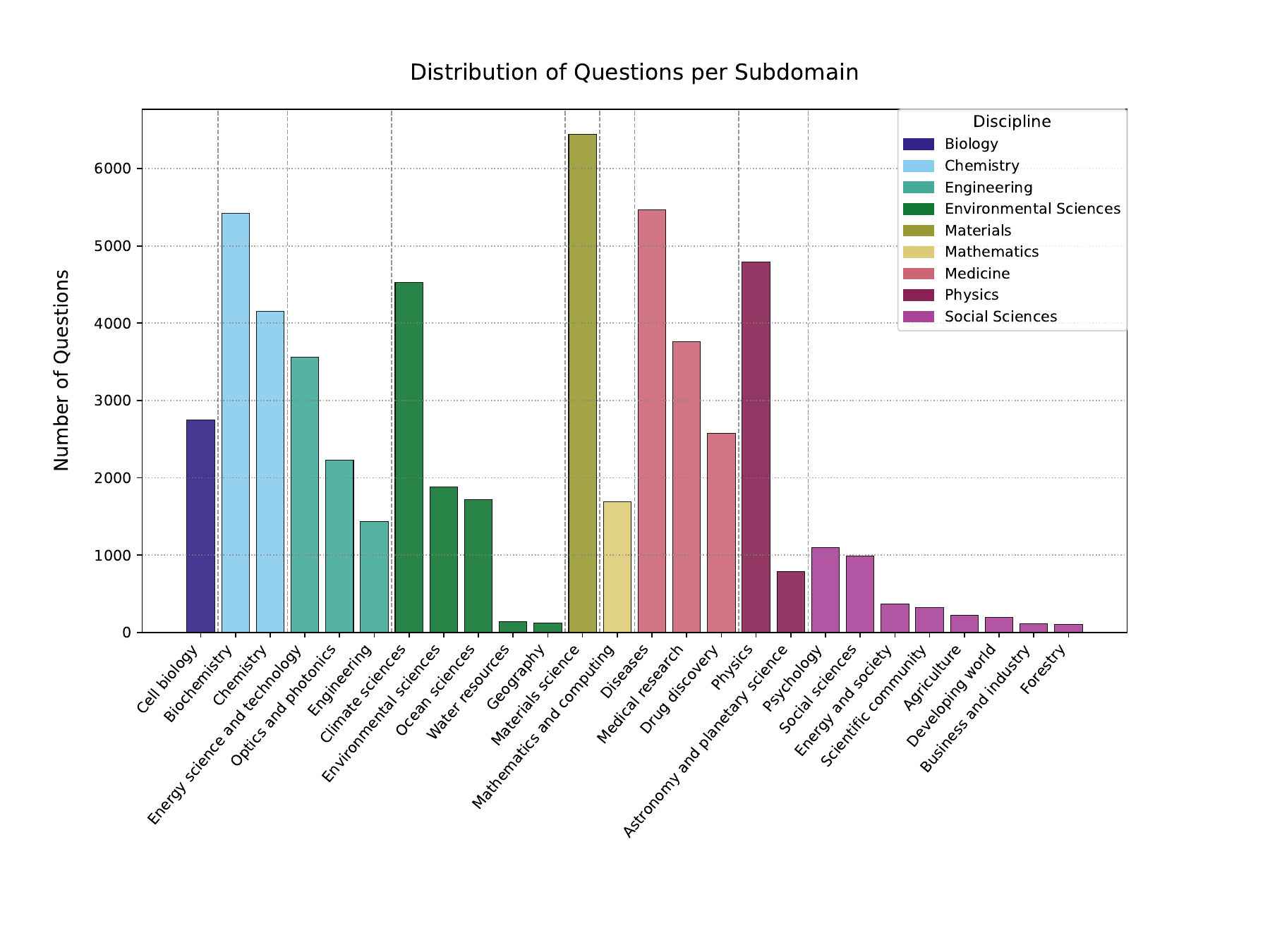}
  \caption{Question distribution across subdomains in WildSci.}
  \label{fig:subdomain_dist}
\end{figure}

We present the distribution of questions across 26 subdomains in Figure~\ref{fig:subdomain_dist}, following the categorization used by Nature Communications. Interdisciplinary areas such as Materials Science and Biochemistry are more prevalent in the journal, leading to a higher number of papers, and consequently more questions from these domains. In contrast, Social Sciences represent a smaller proportion of the journal's content, so we include all available papers from this area in WildSci. The dataset is constructed from papers published prior to April 2024. As new publications become available, our data pipeline can be extended to incorporate the latest research, enabling continuous expansion of both dataset size and knowledge coverage in WildSci.

\begin{figure}
  \centering
  \includegraphics[width=0.6\linewidth]{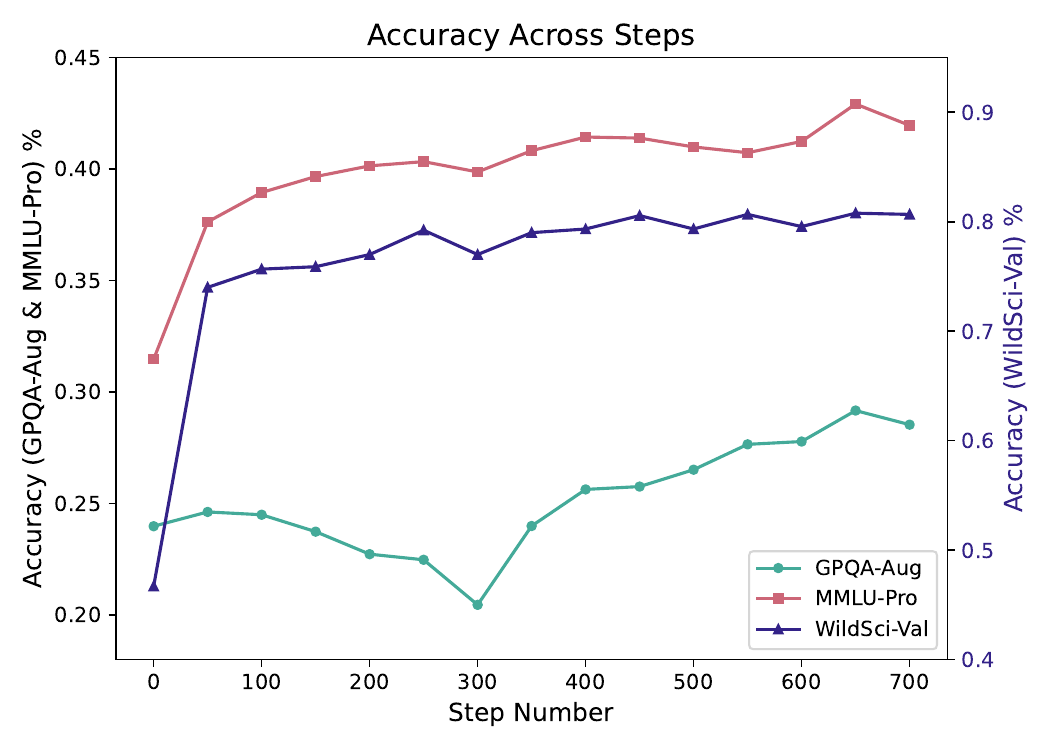}
  \caption{Performance trend on validation and OOD evaluation sets.}
  \label{fig:valid_test}
\end{figure}

\begin{figure}
  \centering
  \includegraphics[width=0.6\linewidth]{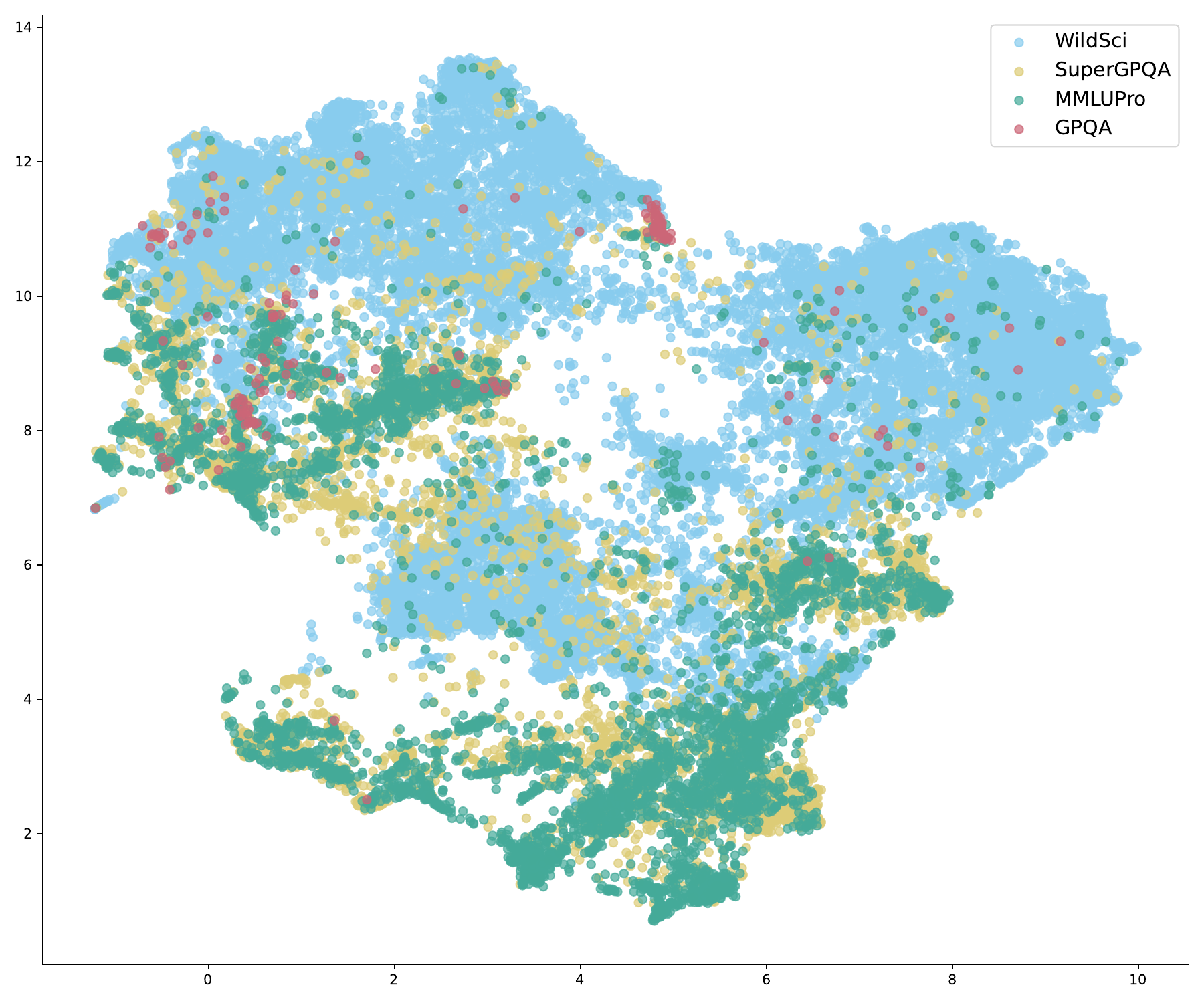}
  \caption{UMAP illustration.}
  \label{fig:umap}
\end{figure}

\section{Different Data Splits}\label{app:data_splits}


Table~\ref{tab:data_splits} compares model performance when trained on different combinations of WildSci subsets. We observe no substantial performance differences across the splits. We hypothesize that this is due to the exploratory and generalization-driven nature of reinforcement learning, which emphasizes domain alignment over data amount. We leave the exploration of how data difficulty and diversity affect RL-based reasoning as a promising direction for future work.

\begin{table}[ht]
\centering
\caption{Performance comparison across different benchmarks. The average is computed across the three public benchmarks GPQA-Aug, SuperGPQA and MMLU-Pro. A.A. means All Aligned, M.A. means Majority Aligned, M.D. means Majority Divergent.}
\resizebox{\linewidth}{!}{%
\begin{tabular}{lccccc}
\toprule
\textbf{Model} & \textbf{WildSci-Val} & \textbf{GPQA-Aug} & \textbf{SuperGPQA} & \textbf{MMLU-Pro} & \textbf{Average} \\
\midrule
Qwen2.5-1.5B-Instruct & 46.70\textsubscript{1.98} & 23.98\textsubscript{0.83} & 18.10\textsubscript{0.63} & 31.47\textsubscript{0.84} & 24.52\textsubscript{0.76} \\
WildSci A.A. & \textbf{80.48}\textsubscript{0.26} & \textbf{28.95}\textsubscript{0.08} & 23.85\textsubscript{0.25} & \textbf{42.54}\textsubscript{0.28} & \textbf{31.78}\textsubscript{0.20} \\
WildSci M.A. & 80.33\textsubscript{0.11} & 25.76\textsubscript{0.44} & \textbf{24.41}\textsubscript{0.06} & 41.95\textsubscript{0.03} & 30.71\textsubscript{0.17} \\

WildSci A.A.+M.A.+M.D. & 80.00\textsubscript{0.59} & 27.48\textsubscript{0.19} & 23.94\textsubscript{0.01} & 41.55\textsubscript{0.17} & 30.99\textsubscript{0.12} \\

\bottomrule
\end{tabular}
}
\label{tab:data_splits}
\end{table}

\section{Domain-Specific Performance}\label{app:per_domain}

We provide a breakdown of subdomain performance in the Table~\ref{tab:subdomain_supergpqa} and Table~\ref{tab:subdomain_mmlupro}. In the model names, `A.A.’ refers to the All Aligned subset, and `M.A.’ refers to the Majority Aligned subset.

\begin{table}[h!]
\centering
\caption{Subdomain performance on SuperGPQA.}
\resizebox{\linewidth}{!}{%
\begin{tabular}{lllllllllllllll}
\toprule
Model&Overall & Agro & Eco & Edu & Eng & Hist & Law & Lite & Mar & Medi & Mili & Phil & Sci & Socio \\
\midrule
1.5B-A.A.&24.12 & 25.98 & 28.98 & 30.37 & 23.47 & 20.62 & 29.42 & 23.21 & 28.94 & 26.39 & 29.76 & 29.40 & 22.66 & 25.87 \\
1.5B-M.A.&24.46 & 24.95 & 28.29 & 32.43 & 23.21 & 21.95 & 30.95 & 23.27 & 29.34 & 25.88 & 32.68 & 29.11 & 23.61 & 27.97 \\
3B-A.A. & 26.39 & 24.33 & 30.59 & 33.26 & 25.82 & 21.81 & 30.94 & 23.21 & 33.33 & 30.02 & 37.56 & 33.43 & 24.88 & 30.07 \\
3B-M.A. & 26.25 & 27.01 & 28.52 & 32.65 & 25.82 & 18.99 & 32.47 & 23.23 & 33.33 & 29.33 & 36.10 & 30.55 & 25.00 & 29.37 \\
\bottomrule
\end{tabular}
}
\label{tab:subdomain_supergpqa}
\end{table}

\begin{table}[h!]
\centering
\caption{Subdomain performance on MMLU-Pro.}
\resizebox{\linewidth}{!}{%
\begin{tabular}{lllllllllllllll}
\toprule
Model& Overall & Bio & Busi & CS & Econ & Health & Math & Other & Psyc & phys & Engin & Hist & law & Philo \\
\midrule
1.5B-A.A. & 42.92 & 64.85 & 50.82 & 46.59 & 54.03 & 38.02 & 57.44 & 35.93 & 53.26 & 40.49 & 29.10 & 34.38 & 22.89 & 36.87 \\
1.5B-M.A. & 41.98 & 64.99 & 48.29 & 40.73 & 53.44 & 36.80 & 55.07 & 35.07 & 53.26 & 39.49 & 30.44 & 32.55 & 24.80 & 35.27 \\
3B-A.A. & 49.48 & 70.71 & 58.56 & 50.73 & 60.43 & 48.04 & 63.43 & 46.54 & 58.77 & 49.04 & 36.02 & 37.01 & 27.16 & 39.08 \\
3B-M.A. & 48.73 & 70.57 & 56.78 & 47.32 & 58.06 & 48.17 & 62.47 & 43.83 & 57.90 & 49.73 & 37.26 & 36.75 & 27.79 & 37.88 \\
\bottomrule
\end{tabular}
}
\label{tab:subdomain_mmlupro}
\end{table}

\section{Post-saturation Generalization}

Beyond the 3B model, we observe similar trends with the 1.5B model.
Figure~\ref{fig:valid_test} illustrates performance during training on the Majority Aligned subset. While validation accuracy fluctuates after step 400, accuracy on GPQA-Aug and MMLU-Pro continues to improve, indicating post-saturation generalization. We leave investigation of the underlying causes to future work.

\section{Distribution Illustration}

To compare the coverage of WildSci with existing science benchmarks, we embed the question texts (excluding answer options) using a pretrained sentence embedding model and apply UMAP~\citep{DBLP:journals/corr/abs-1802-03426} for dimensionality reduction. Figure~\ref{fig:umap} presents a 2D visualization of questions from different datasets. We randomly sample 5,000 questions from SuperGPQA, MMLU-Pro respectively and 20,000 questions from WildSci.
In the figure, SuperGPQA and MMLU-Pro share similar distributions, WildSci occupies regions underrepresented by these benchmarks. The partial mismatch in distribution suggests that existing benchmarks serve as effective out-of-distribution (OOD) evaluations for WildSci.

\section{Prompts}\label{app:prompts}

In this section, we show the prompts we use in our data creation pipeline. During QA generation, we randomly sample two QAs from the GPQA-Extended set, excluding those in the Diamond subset, to serve as JSON examples.

\newpage

\definecolor{mydynamiccolor}{HTML}{117733}

\begin{tcolorbox}[colback=mydynamiccolor!20, colframe=mydynamiccolor, title=\textbf{QA Generation}]
You are an expert research scientist in [[DOMAIN]]. Your task is to extract or create three challenging and difficult, self-contained question–answer pairs from the provided academic paper. The QAs will be used as exam questions for PhD students and must be clear, extremely challenging, and context-independent, i.e., understandable on its own without referring back to the original paper.\\

Each QA pair must include:\\

1. Question:\\
- A clear, difficult, and standalone question.\\
- The question must include sufficient background information or context so that one can fully understand and attempt it without referring to the original paper.\\
- Define any abbreviations, notation, or domain-specific terminology used.\\
- DO NOT use phrases like “according to the paper” or “the proposed method.”\\
- The question should be complex enough to require deep understanding of the subject.\\
- It should engage **advanced reasoning**, such as: Conceptual analysis, Theoretical or mathematical derivations, Methodological design, Causal reasoning or hypothesis testing, etc.\\

2. Options:\\
- Provide four answer choices.\\
- Only one option should be correct.\\
- The three incorrect options should be plausible but clearly wrong upon careful reasoning, ideally derived by subtly altering the logic or assumptions behind the correct answer.\\

3. Answer:\\
The correct option letter.\\

4. Rationale:\\
- A detailed explanation of why the correct answer is correct and why each incorrect option is wrong.\\
- DO NOT reference the original paper in any part of the rationale.\\
- If calculations are required, include the full step-by-step process.\\

Important:\\
- Questions must be self-contained, including any necessary context or definitions.\\
- Do not reference the original paper in any part of the question, options, or rationale.\\
- Aim for PhD-level difficulty, testing understanding of key technical ideas.\\
- Ensure that only one option is unambiguously correct.\\

Format your QA pairs in the following JSON format, here are examples:
[[JSON example]]
\end{tcolorbox}

\begin{tcolorbox}[colback=mydynamiccolor!20, colframe=mydynamiccolor, title=\textbf{Refinement}]

You are an expert research scientist in [[DOMAIN]].\\
Your task is to refine each provided multiple-choice question to increase its difficulty and test deeper reasoning appropriate for PhD-level understanding. Each item includes a question, answer options, and a solution. Use the provided answers and solution for reference, they may be incorrect.\\

Refinement Goals:\\
1. Expand Options with Subtle Variations\\
Rewrite the answer options to include 10 choices labeled A–J.\\
All options should seem plausible to someone with partial knowledge, but only one should be fully correct. Introduce subtle numerical or conceptual variations among options.\\
2. Remove Surface-Level Hints\\
Eliminate obvious formulas, definitions, or axioms from the question.\\
Assume the solver must recall or derive them independently. You may include these concepts in the solution rationale, but not in the question.\\
3. Increase Reasoning Depth\\
Replace direct or few-step problems with multi-step (>3) or causal reasoning.\\
Use reasoning chains (e.g., X leads to Y, which leads to Z, which explains W) or require intermediate inferences without giving all variables.\\
4. Rephrase to Introduce Diversity\\
Use varied question formats: causal, hypothetical, comparative, inferential, conditional, etc.\\
Maintain clarity and scientific rigor while diversifying expression.\\

Important:\\
- Each refined question should challenge deep understanding of core technical concepts, appropriate for advanced graduate-level assessment.\\
- The refined question must remain solvable using information provided or generally assumed background knowledge in the domain. Avoid ambiguity or underspecified problems.\\
- Ensure that only one option is unambiguously correct.\\
- If the original question is poorly written, ambiguous, or lacks depth, you may create a new question based on the same underlying concept or topic reflected in the provided QA.\\

Please only output the refined QA in the following JSON format:
[[JSON example]]

\end{tcolorbox}

\newpage